\documentclass[journal]{IEEEtran}

\usepackage{cite}

\usepackage[pdftex]{graphicx}
\ifCLASSINFOpdf
\else
\fi
\usepackage[cmex10]{amsmath}
\usepackage{algorithm}
\usepackage{algorithmic}

\usepackage{color}

\usepackage{multicol}
\usepackage{multirow}

\usepackage[caption=false,font=footnotesize]{subfig}

\begin{document}
%
\title{Cascaded Subpatch Networks for Effective CNNs}
%
%
%

\author{Xiaoheng~Jiang,
        Yanwei~Pang,
        Manli~Sun,
        and~Xuelong~Li
\thanks{Y. Pang, X. Jiang, and M. Sun are with the School of Electronic Information Enginnering, Tianjin University, Tianjin 300072, China. (E-mail: \{pyw, jiangxiaoheng, sml\}@tju.edu.cn).}
\thanks{X. Li is with the Center for OPTical IMagery Analysis and Learning (OPTIMAL), State Key Laboratory of Transient Optics and Photonics, Xi'an Institute of Optics and Precision Mechanics, Chinese Academy of Sciences, Xi'an 710119, Shaanxi, P. R. China. E-mail: xuelong{\_}li@opt.ac.cn}
}

\maketitle

\begin{abstract}
Conventional Convolutional Neural Networks (CNNs) use either a linear or non-linear filter to extract features from an image patch (region) of spatial size $ H\times W $  (Typically, $ H $ is small and is equal to $ W$, e.g., $ H $  is 5 or 7).  Generally, the size of the filter is equal to the size $ H\times W $  of the input patch. We argue that the representation ability of equal-size strategy is not strong enough. To overcome the drawback, we propose to use subpatch filter whose spatial size $ h\times w $  is smaller than $ H\times W $. The proposed subpatch filter consists of two subsequent filters. The first one is a linear filter of spatial size $ h\times w $ and is aimed at extracting features from spatial domain. The second one is of spatial size $ 1\times 1 $  and is used for strengthening the connection between different input feature channels and for reducing the number of parameters. The subpatch filter convolves with the input patch and the resulting network is called a subpatch network. Taking the output of one subpatch network as input, we further repeat constructing subpatch networks until the output contains only one neuron in spatial domain. These subpatch networks form a new network called Cascaded Subpatch Network (CSNet). The feature layer generated by CSNet is called csconv layer. For the  whole input image, we construct a deep neural network by stacking a sequence of csconv layers. Experimental results on four benchmark datasets demonstrate the effectiveness and compactness of the proposed CSNet. For example, our CSNet reaches a test error of $ 5.68\% $ on the CIFAR10 dataset without model averaging. To the best of our knowledge, this is the best result ever obtained on the CIFAR10 dataset.
\end{abstract}


%
\IEEEpeerreviewmaketitle

\section{Introduction}
%
%
%
%
\IEEEPARstart{C}{ONVOLUTIONAL} neural networks (CNNs) \cite{Jia_Caffe_CoRR2014,Long_FullyConvolu_CVPR2015} have achieved a great success in the field of computer vision, including image classification  \cite{He_SpatialPyramid_ECCV2014,Krizhevsky_Imagenet_NIPS2012,Simonyan_VeryDeep_CoRR2014,Agrawal_AnalyzingPerformance_ECCV2014,Ji_Relvevance_TIP2015,Pang_Learning_TNNLS2014} and object detection \cite{Girshick_FastRcnn_ICCV2015,Girshick_CVPR2014,Ren_FasterRCNN_NIPS2015,Jiang_SpeedUp_Neuro2015,Pang_Sampling_CoRR2015}.  The underlying reason lies in the fact that CNN is able to learn a hierarchy of features \cite{Chang_DeepShallow_TNNLS2015,Bengio_PAMI2013,Gong_AMultiObjective_TNNLS2015,Wu_Leveraging_CVPR2014}   that can represent objects in different levels. Low-level features denote some visual features such as edges, dots, and textures, whereas high-level features represent objects in a semantic way. Low-level features are shared by all objects while high-level features are of high discriminability. High-level features are learned progressively from low-level features. All these features are in fact learned through a series of linear and non-linear transformations which are the primary elements of CNNs.

Typically, CNN consists of several computational building blocks: convolution, activation, and pooling. They work together to fulfill the task of feature extraction and transformation. Convolution takes inner product of the linear filter and the local region of input channel. Activation imposes a non-linear transformation on the convolutional results. Pooling gathers the responses of a given region. Among these three building blocks, convolutional block plays the most important role in CNN. It controls the number of feature maps (i.e., width of CNN) and the number of layers (i.e., depth of CNN). The width and depth determine the capacity of CNN. The size of neural network is a double-edged sword. On the one side, large size means large capacity. Large capacity makes it possible for deep networks to learn rich features which are essentially important for task of recognizing tens or even thousands of object categories. On the other side, large size typically means a larger number of parameters, which makes the enlarged network more prone to over-fitting especially when the number of labelled samples in the training set is limited. What is more, the main drawback of large network is the dramatically increased consumption of computational resources.

To construct a compact and powerful network, we propose a novel type of convolutional filters. Given a local patch, traditional CNNs typically use a convolutional filter which is the same size as the patch to extract features. We argue that this level of abstraction is not strong enough to generate robust features. Multi-Layer Perceptron (MLP) \cite{Lin_NIN_CoRR} can be used to impose more complex transformation. However, it is still not complex enough to represent the input data which lies on a highly non-linear manifold. Therefore, in this paper, we propose to use cascaded subpatch filters to bring in much more complex structures to abstract the local patch within the receptive field. One subpatch filter contains two subsequent convolutional filters. The first one abstracts subpatches of the input patch. The second one is to fully connect all the output channels of the first one. Taking the convolutional output of previous subpatch filter as input, we repeat constructing new subpatch filters until the final output contains only one neuron in spatial domain. This results in the Cascaded Subpatch Network (CSNet). CSNet can be used to replace conventional convolutional layer to extract more complex and more robust features. We call the resulting layer a csconv layer. A deep neural network can be obtained by stacking multiple csconv layers. For clarity, in the rest of this paper, the overall deep network containing multiple csconv layers is called a CSNet. 

The goal of the proposed method is to construct a more effective structure to abstract the local patch. Instead of designing a CNN that is too wide (i.e., too many feature maps in one layer) or too lengthy (i.e., too many layers), we present a novel neural network which is compact, yet powerful. Specifically, the contributions and merits of this paper are summarized as follows.
\begin{enumerate}

\item We gain new insight into the convolutional block of CNN. When abstracting one local patch of size $ H\times W $, we propose to use cascaded subpatch filters to replace conventional convolutional filter. 

\item Subpatch filter consists of an $ h\times w $  linear filter followed by a $ 1\times 1 $  filter. Its purpose is to impose a complex transformation on subpatches of the input patch while reducing the number of parameters.

\item Cascaded subpatch filters contain a sequence of subpatch filters and they together reach the goal of generating a more complex and more robust abstraction of the local patch. The cascaded subpatch filters can be regarded as one new convolutional kernel structure called csconv filter. Csconv filter abstracts local patch much better than conventional filter.

\item Csconv filter is a flexible structure which can be constructed using a group of different subpatch filters according to size of the local region and the demanding number of parameters. 

\item We build several CSNets with different number of parameters to deal with different tasks. And our CSNets achieve the state-of-the-art performance on four widely used benchmark image classification datasets. 
\end{enumerate}

This paper is organized as follows. Section II reviews the related work. Section III presents the proposed CSNet method. The experimental results are given in Section IV. Finally, Section V concludes this paper.

\section{Related Work}
Since the great success of AlexNet \cite{Krizhevsky_Imagenet_NIPS2012}   on the ImageNet Large Scale Visual Recognition Challenge (ILSCRC-2010), a number of attempts have been made to improve the architectures of CNN in order to achieve better accuracy. We divide these methods into the following three categories.

(1) \textbf{Parameter adjusting}. Some researchers paid attention to the parameters of CNN, such as the sizes of convolutional filters, the strides of filters, the number of feature channels in each layer, and the number of convolutional layers. They tried to adjust the parameters to improve the performance of CNN through exhaustive experiments. Zeiler and Fergus \cite{Zeiler_VisualingUnderstanding_ECCV2014}  visualized the trained CNN model and found that large filter size and large stride of the first convolutional layer could cause aliasing artifacts. Therefore, they used smaller receptive window size and smaller stride. Sermanet \textit{et al.} \cite{Sermanet_Overfeat_CoRR2013}  utilized smaller strides in the first convolution, larger number of feature maps, and larger number of layers. They achieved better results than the AlexNet. The VGG network \cite{Simonyan_VeryDeep_CoRR2014}  pushes the depth of CNN to up to 19 convolutional layers by using very small convolutional filters and gains a significant improvement. These above efforts can be viewed as preliminary explorations on how to construct networks with better performance.

(2) \textbf{Structure designing}. Another line of improvements go further into the designing of new CNN structures. Network in Network (NIN) \cite{Lin_NIN_CoRR}  utilizes shallow Multi-Layer Perception (MLP) to increase the representational power of neural networks. MLP is a more complex structure which consists of multiple fully connected layers. Conventional linear convolution and MLP together result in a new convolutional structure called mlpconv. Mlpconv can be easily implemented by stacking additional $ 1\times 1 $ convolutional layers on conventional convolutional layer. These $ 1\times 1 $ convolutions actually enhance the connection between different feature channels. Therefore, mlpconv is able to abstract local regions much more effectively than conventional convolution. Szegedy \textit{et al.} \cite{Szegedy_GoingDeeper_CVPR2015}  constructed a 22-layer GoogleNet by stacking dozens of Inception modules. Each Inception module contains a group of convolutional filters of different sizes which aim at capturing information of multiple scales. However, such an Inception module is too wide to be efficiently used in a very deep network. To overcome the disaster of having too many parameters, GoogleNet takes advantage of $ 1\times 1 $ convolutions as dimension reduction modules to remove computational bottlenecks. As a result, it allows for increasing not just the depth but also the width of the GoogleNet without significant increasing in parameters. 

(3) \textbf{Deeper and wider networks}. Since increasing the size of CNN is the most straightforward way to improve their performance, researchers went even further into designing much deeper networks which are up to hundreds or even thousands of layers. Highway Networks \cite{Srivastava_TrainingVeryDeep_NIPS2015} make it possible to train very deep networks even with hundreds of layers by using adaptive gating units to regulate the information flow. More importantly, Highway Networks are able to train deeper networks without sacrificing generalization ability. The 32-layer highway network presented in \cite{Srivastava_TrainingVeryDeep_NIPS2015} achieved the state-of-the-art performance on the CIFAR10 \cite{Krizhevsky_LearningMultiple_Master2009} dataset. Based on Highway Networks, He \textit{et al.} \cite{He_DeepResidual_CoRR2015} recently presented a residual learning framework to effectively train networks which are substantially deeper than ever used. They constructed residual nets (ResNest) with a depth of up to 152 layers and evaluated them on the ImageNet dataset. They also presented ResNets with 100 and 1000 layers and evaluated them on the CIFAR10 dataset. They argued that the depth of representations is of great importance for many visual recognition tasks.  In addition to the depth of CNN, the width of CNN is also very important. Shao et al. \cite{Shao_LearningDeep_TNNLS2014}  pointed out that the combination of multicolumn deep neural networks could enhance the robustness. Instead of simply averaging the outputs of multicolumn predictions, they learned a compact representation from multicolumn deep neural networks by embedding the features of all the penultimate layers into a multispectral space. The resulting features are then used for classification. Their multispectral neural networks (MSNN) in fact make use of the complementary information captured by different neural networks. Since the MSNN has to use multiple networks that do not share parameters, the computation increases with the number of networks.

We agree that both width and depth of networks are important for the tasks of visual recognition. However, larger capacity does not guarantee higher accuracy. Given a dataset with limited samples, when a network reaches its peak performance, it is difficult to further improve performance by simply adding more feature maps or stacking more convolutional layers. This means that the discriminability of networks does not increase infinitely with the size of networks. The performance comparison with ResNet110 \cite{He_DeepResidual_CoRR2015}  and ResNet1202 \cite{He_DeepResidual_CoRR2015}  on the CIFAR10 dataset supports our viewpoint. ResNet110 is a 110-layer CNN with 1.7M parameters, and ResNet1202 is a 1202-layer CNN with up to 19.4M parameters. However, ResNet110 achieves a test error of $ 6.43\% $ whereas ResNet1202 achieves a test error of $ 7.93\% $. That is, the classification ability of deep neural network may suffer from excessive parameters. Therefore, it is important to explore new methods to learn features in a more effective way.

\section{Proposed Method}
This paper is aimed at using subpatch filters to construct compact and powerful CNNs. One of the characteristics of the proposed method is that the size of subpatch filter is smaller than that of the patch to be presented. In our method, cascaded subpatch filters are used to represent a patch. We take the cascaded subpatch filters as a whole and call it csconv filter. Applying the csconv filters layer by layer results in a deep CNN which we call CSNet. In this section, we first introduce the subpatch filter. Next, we describe cascaded subpatch filters. Then, the CSNet is presented. Finally, analysis of the computational complex of CSNet is given.

\begin{figure}[!t]
\centering
\includegraphics[scale=0.6]{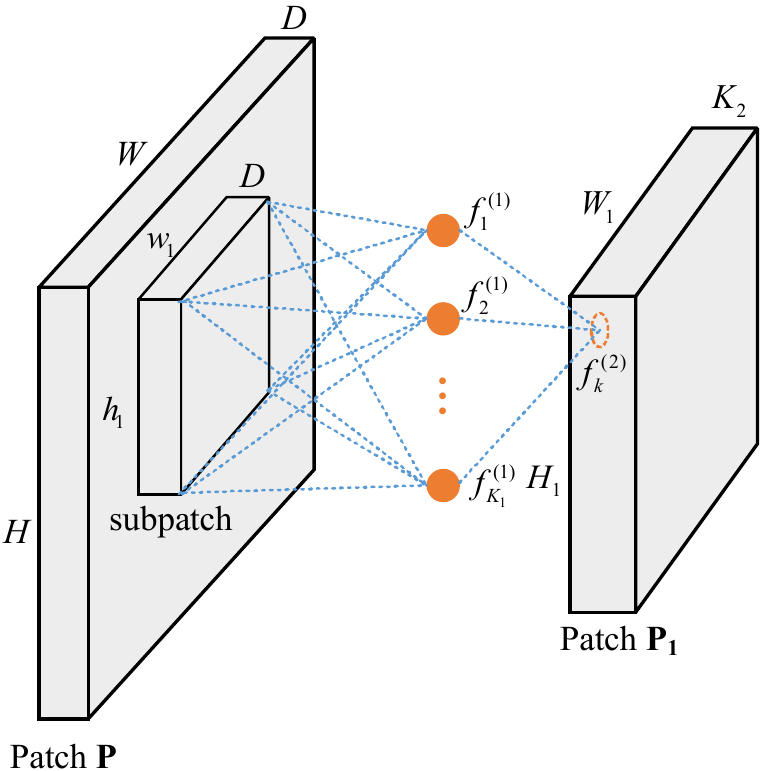}
\caption{A subpatch filter  $ \textbf{w}_{k}=(\textbf{w}_{k}^{(1)},\textbf{w}_{k}^{(2)})$ of size $ (h_{1}\times w_{1},1\times 1) $ consists of a $ h_{1}\times w_{1} $ filter $ \textbf{w}_{k}^{(1)} $  and a $ 1\times 1 $ filter $ \textbf{w}_{k}^{(2)} $ . The input is a patch $  \textbf{P}  $ of size $ H\times W$ and the convolutional output is a patch $ \textbf{P}_{1} $ of size $ H_{1}\times W_{1} $ with $ H_{1}= (H-h_{1}+1)$ and $ W_{1}=(W-w_{1}+1) $.}
\label{Fig1}
\end{figure}

\begin{figure}[!t]
\centering
\subfloat[A conventional filter.]{\includegraphics[scale=0.5]{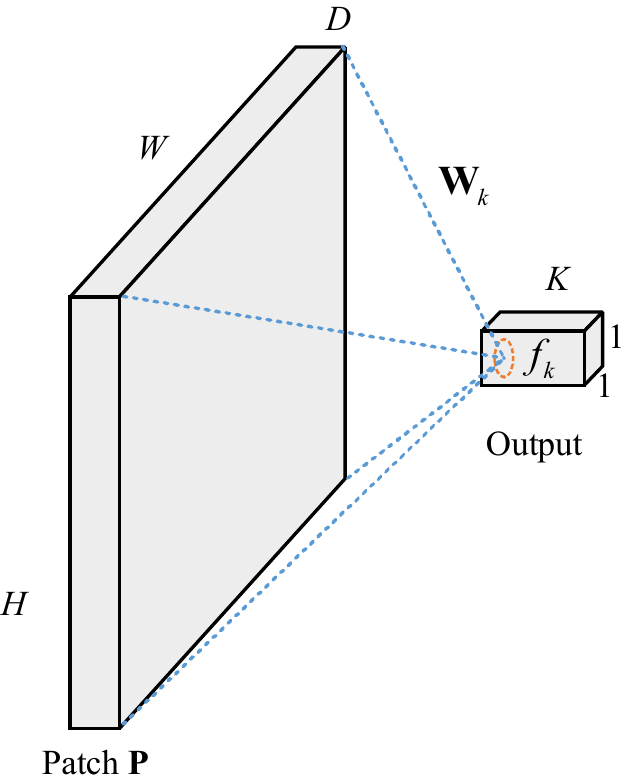}
\label{Fig2(a)}}
\hfil
\subfloat[An $ n $-stage csconv filter]{\includegraphics[scale=0.5]{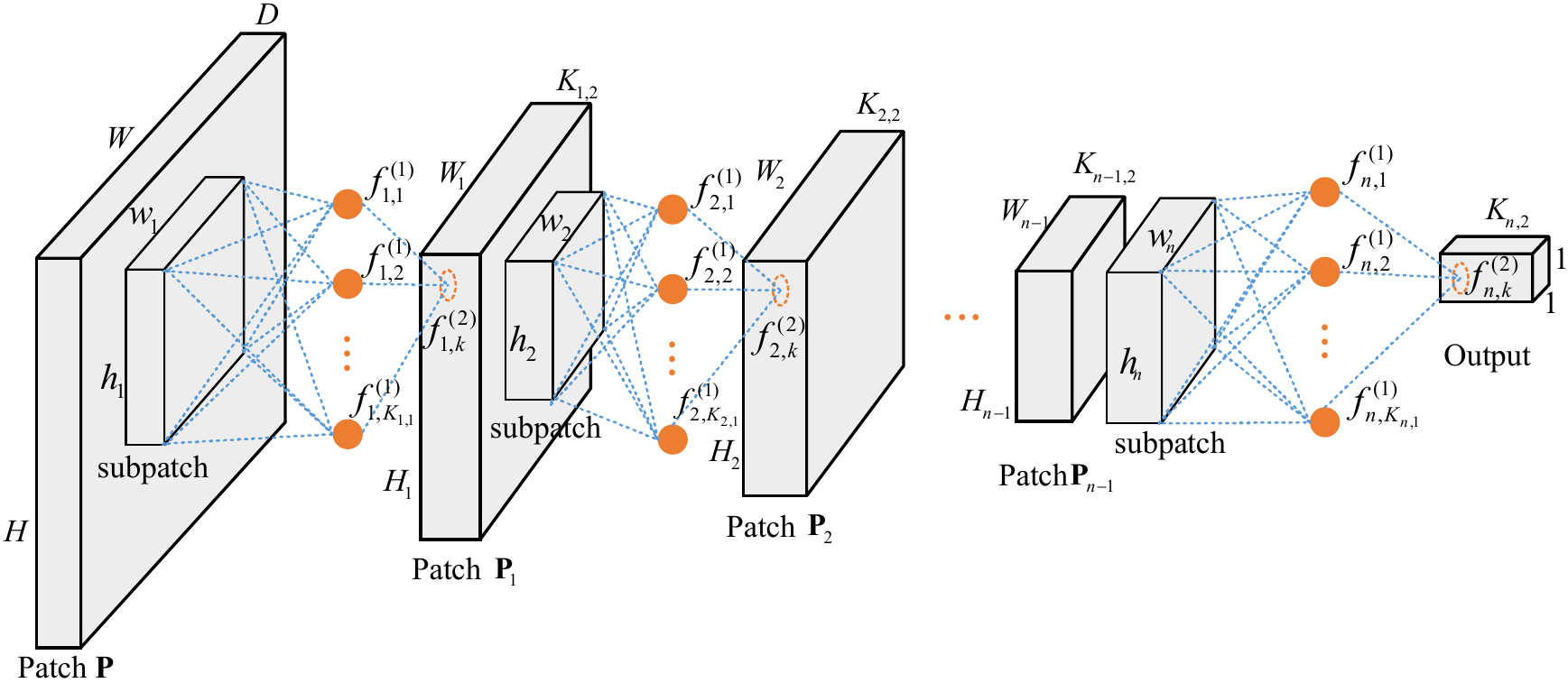}
\label{Fig2(b)}}
\caption{Comparison between conventional filter and the proposed csconv filter. (a) A conventional convolutional filter $ \textbf{W}_{k} $ that is the same size as the input patch of size $ H\times W$. (b) An $ n $-stage csconv filter $ [(h_{1}\times w_{1},1\times 1),(h_{2}\times w_{2},1\times 1),...,(h_{n}\times w_{n},1\times 1)] $. The input is a patch $  \textbf{P}  $ of size $ H\times W$ and the final output is of size $ 1\times 1 $.}
\label{Fig2}
\end{figure}

\subsection{Subpatch Filter}
The task is to represent an input patch $ \textbf{P} \in \mathcal{R}^{H\times W \times D} $ where $ H\times W $ stands for the spatial size and $ D $ is the number of channels. Throughout this paper, the spatial size is used to express the patch size. By vectoring the three-order tensor, $ \textbf{P} $ can be expressed as an $ H\times W \times D $ dimensional column vector $ \textbf{X}\in \mathcal{R}^{{(H\times W \times D)}\times 1} $. Conventional convolution uses a linear filter $ \textbf{W}_{k}\in \mathcal{R}^{{(H\times W \times D)}\times 1} $ whose size is the same as the patch \textbf{X}. The conventional convolution can be computed by inner product
\begin{equation}
\label{eq1}
f_k =  {\textbf{W}_k}^T\textbf{X}\in \mathcal{R}^{1}, k=1,2,...,K
\end{equation}
where $ K $ is the number of output channels. The convolution converts the patch of spatial size $ H\times W $ into a scalar $ f_{k} $. For the sake of notation consistence, we use $ 1\times 1 $ to represent the size of feature $ f_{k} $. Fig.~\ref{Fig2}(a) shows the conventional convolution.

To make the feature representation more effective, we propose to utilize cascaded subpatch filters to transform the patch from size $ H\times W $ to $ 1\times 1 $. Let  $ \textbf{x}\in \mathcal{R}^{{(h\times w \times D)}\times 1} $ is subpatch of \textbf{X} with $ h<H $ and $ w<W $. The number of overlapping subpatches of size $ h\times w $ in the patch of size $ H\times W $ is $ N=(H-h+1)\times (W-w+1) $. A subpatch filter $ \textbf{w}_{k} $ consists of two subsequent filters with the size of the first filter $ \textbf{w}_{k}^{(1)}\in \mathcal{R}^{{(h\times w \times D)}\times 1} $ being $ h\times w $ and the size of the second filter $ \textbf{w}_{k}^{(2)}\in \mathcal{R}^{{(1\times 1 \times D)}\times 1} $ being $ 1\times 1 $. To explicitly show that a subpatch filter $ \textbf{w}_{k} $ is composed of two basic filters $ \textbf{w}_{k}^{(1)} $ and $ \textbf{w}_{k}^{(2)} $, we denote the subpatch filter $ \textbf{w}_{k} $ by $ \textbf{w}_{k}=(\textbf{w}_{k}^{(1)},\textbf{w}_{k}^{(2)})$ and denote the size of subpatch filter by 
$ (\textit{h}\times \textit{w},1\times 1) $. We call the second filter $ \textbf{w}_{k}^{(2)} $ channel filter because its function is to fully connect different channels. We call the first filter $ \textbf{w}_{k}^{(1)} $ spatial filter because its size is larger than $ 1\times 1 $ and its role is to extract features from both spatial and channel domain.

The inner product between the spatial filter $ \textbf{w}_{k}^{(1)} $ and the subpatch \textbf{x} is 
\begin{equation}
\label{eq2}
f_k^{(1)} =  {(\textbf{w}_k^{(1)})}^T\textbf{x}\in \mathcal{R}^{1}, k=1,2,...,K_{1}
\end{equation}
where $ K_{1} $ is the number of output channels. Express the output of the spatial filter as a $ K_{1} $-dimensional feature vector $ \textbf{f}^{(1)}={(f_1^{(1)},f_2^{(1)},...,f_{K_{1}}^{(1)})} \in \mathcal R^{K_{1}}$. Taking the feature vector as the input of the channel filter $ \textbf{w}_{k}^{(2)} $, the second inner product is obtained by
\begin{equation}
\label{eq3}
f_k^{(2)} =  {(\textbf{w}_k^{(2)})}^T\textbf{f}^{(1)}\in \mathcal{R}^{1}, k=1,2,...,K_{2}
\end{equation}
where $ K_{2} $ is the number of output channels.

Eq.~\ref{eq2} and Eq.~\ref{eq3} are only involved in one subpatch. There are $ N=(H-h+1)\times (W-w+1) $ subpatches. So we apply Eq.~\ref{eq2} and Eq.~\ref{eq3} on all the $ N $ subpatches. That is, the subpatch filter $ \textbf{w}_{k}  $ of size $ (h\times w, 1\times 1) $ convolves with the input patch $ \textbf{P} \in \mathcal{R}^{H\times W \times D} $. The convolution is conducted without zero-padding. Consequently, the output $ \textbf{P}_{1} $ of the convolution with subpatch filter $ \textbf{w}_{k} $ is a patch of size $ H_{1}\times W_{1} $ where $ H_{1} $ is $ (H-h+1) $ and $ W_{1} $ is $ (W-w+1) $. Fig.~\ref{Fig1} demonstrates one subpatch filter of size $(h_{1}\times w_{1},1\times 1)  $.

\begin{figure}[!t]
\centering
\subfloat[A conventional filter]{\includegraphics[scale=0.5]{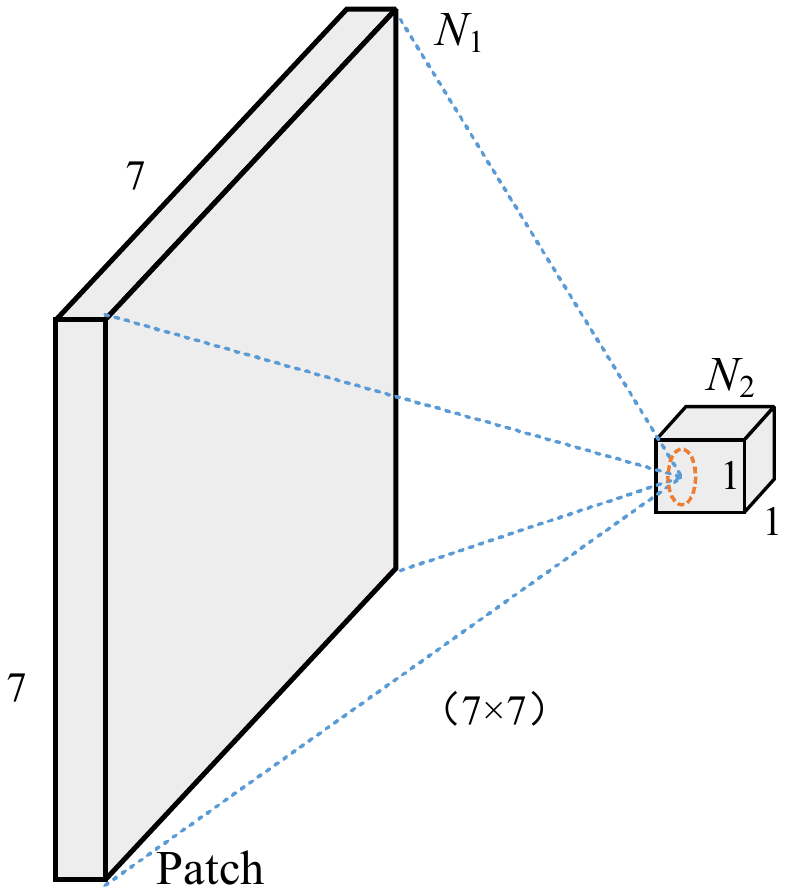}
\label{Fig3(a)}}
\hfil
\subfloat[A two-stage csconv filter]{\includegraphics[scale=0.5]{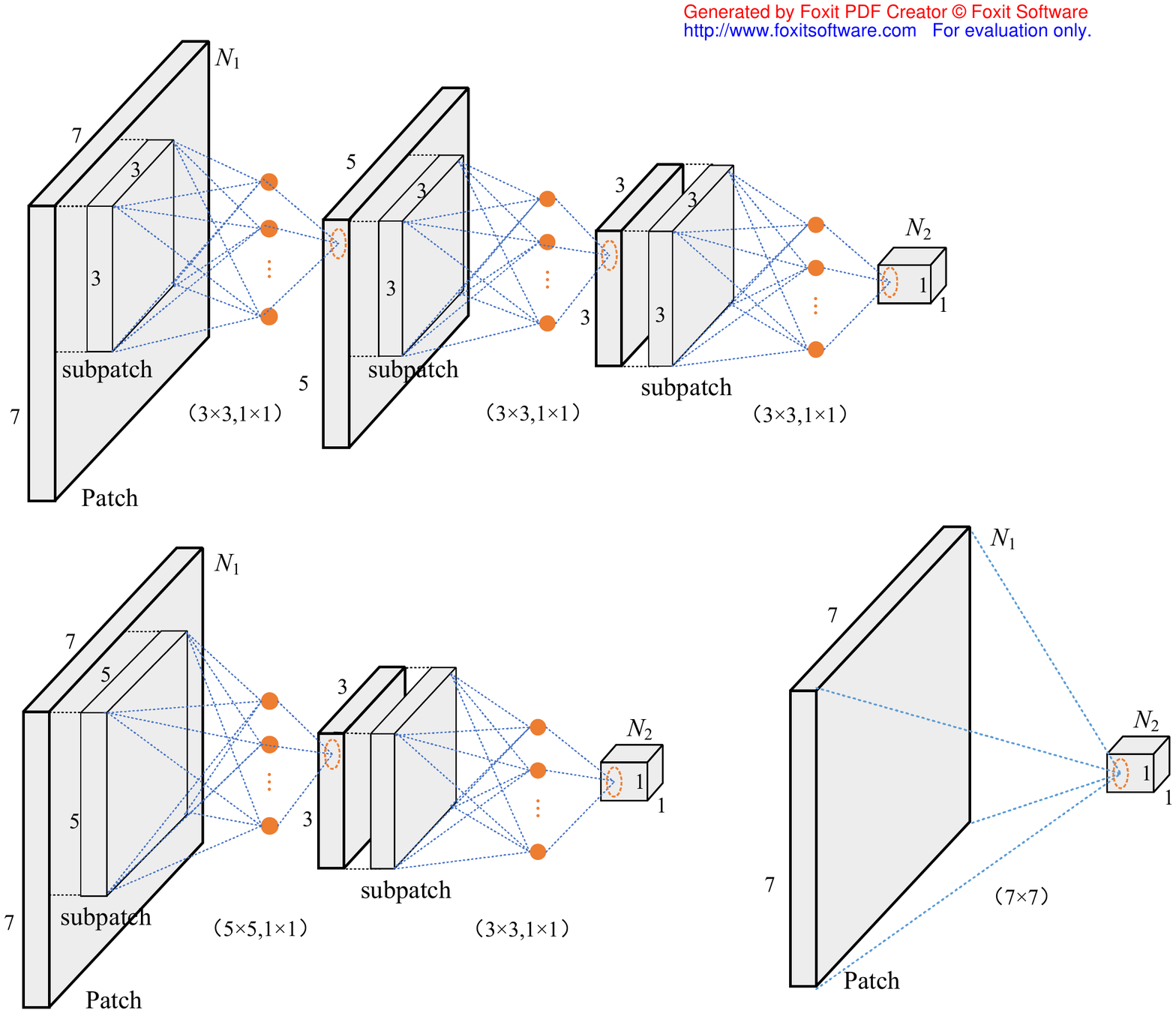}
\label{Fig3(b)}}
\hfil
\subfloat[A three-stage csconv filter]{\includegraphics[scale=0.5]{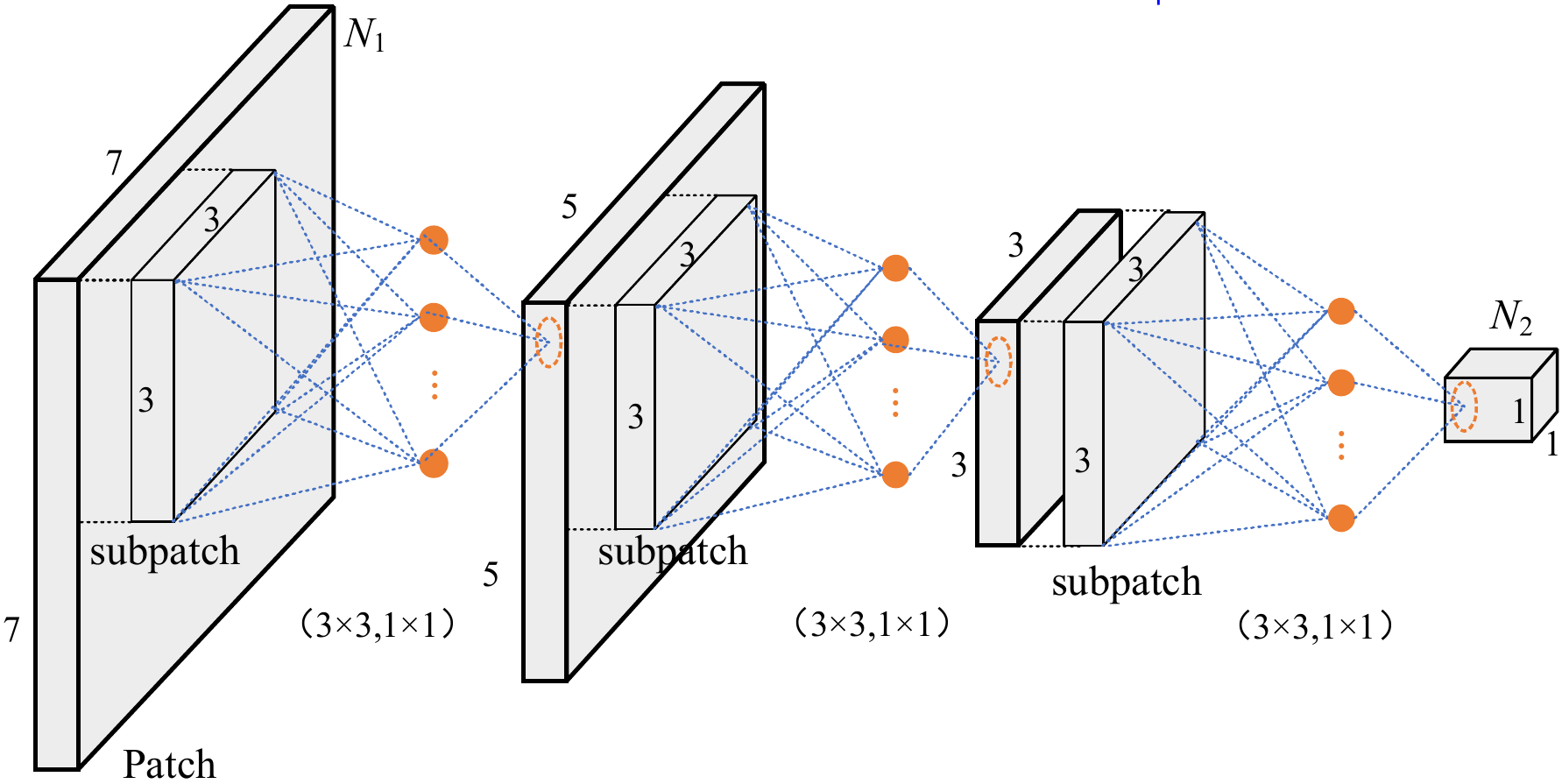}
\label{Fig3(c)}}
\caption{Abstrcating the input patch of size $ 7\times 7 $ using different filters. (a) Convolution with a conventional filter of size $ 7\times 7 $, directly generating an output of size $ 1\times 1 $. (b) Convolution with a two-stage csconv filter $ [(5\times 5,1\times 1),(3\times 3,1\times 1) ]$, generating one intermediate feature layer (consisting of a number of feature channels) of size $ 3\times 3 $. (c) Convolution with a three-stage csconv filter $ 3*(3\times 3,1\times 1) $, generating two intermediate feature layers of sizes $ 5\times 5 $ and $ 3\times 3 $, respectively. }
\label{Fig3}
\end{figure}

\subsection{Generating a Csconv Filter by Cascaded Subpatch Filters}
In the previous section, we explicitly denote a subpatch filter by $ \textbf{w}_{k}=(\textbf{w}_{k}^{(1)},\textbf{w}_{k}^{(2)})$ where $ k $ indexes channels. When there are several subpatch filters of different sizes, for the sake of clarity, we denote the $ i $-th subpatch filter of size $ (h_{1}\times w_{1},1\times 1) $ by $ \textbf{w}_{i}=(\textbf{w}_{i,k}^{(1)},\textbf{w}_{i,k}^{(2)})$. Fig.~\ref{Fig1} shows that convolving a subpatch filter $ \textbf{w}_{1} $ of size $ (h_{1}\times w_{1},1\times 1) $ within the input patch of size $ H\times W $ results in an output patch $\textbf{ P}_{1} $ of size $ H_{1}\times W_{1}= (H-h_{1}+1)\times (W-w_{1}+1) $. But our goal is to output a $ 1\times 1 $ patch to represent the input \textbf{P}. This goal can be arrived at by convolving the output patch $\textbf{ P}_{1} $ with another subpatch filter $ \textbf{w}_{2} $ of size $ (h_{2}\times w_{2},1\times 1) $ with $ h_{2}\leq h_{1} $ and $ w_{2}\leq w_{1} $. The size $ H_{2}\times W_{2} $ of the output of $ \textbf{w}_{2} $ is  $ (H_{1}-h_{2}+1)\times (W_{1}-w_{2}+1) $. It can be noted that $ H_{2}<H_{1} $ and $ W_{2}<W_{1} $. That is, once a subpatch filter is used, the size of the output patch is decreased. The subpatch filters are subsequently used until the output is of size $ 1\times 1 $. Specially, th sizes of spatial filters can be expressed as:
\begin{equation}
\label{eq4}
\left\{ \!{\begin{array}{l}
H_{1} = H - h_1 + 1,W_{1} = W - w_1 + 1; \\ 
H_{2} = H_{1} - h_2 + 1,W_{2} = W_{1} - w_2 + 1; \\ 
... \\ 
H_{n \!-\! 1} \!=\! H_{n \!-\! 2} \!- \!h_{n\! -\! 1} \!+ \!1, W_{n\! - \!1}\! =\! W_{n \!- \!2}\! -\! w_{n\! - \!1} \!+ \!1; \\ 
H_n \!=\! H_{n \!-\! 1} \!-\! h_n \!+\! 1 \!=\! 1,W_{n}\! =\! W_{n \!-\! 1}\! -\! w_n \!+\! 1 \!= \!1. \\ 
\end{array}} \right.
\end{equation}
It is noted that the size  $ H_{n \!-\! 1}\times W_{n \!-\! 1}$ of penultimate output patch is the same as that $ h_{n}\times w_{n}$ of the spatial filter of the last subpatch filter.

Suppose that $ n $ subpatch filters are finally used to obtain a $ 1\times 1 $ output patch. We denote the cascaded $ n $ subpatch filters by $ [(h_{1}\times w_{1},1\times 1),(h_{2}\times w_{2},1\times 1),...,(h_{n}\times w_{n},1\times 1)] $. If all the $ n $ subpatch filters have the same size $ h\times w $ (i.e., $ h_{1}=h_{2}=...h, w_{1}=w_{2}=...w $), then the cascaded filter can be denoted by $ n*(h\times w,1\times 1) $. We call the $ n $ cascaded subpatch filters $ n $-stage csconv filter. Fig.~\ref{Fig2}(b) demonstrates an $ n $-stage csconv filter. 

Given a local patch, different filters can be used to deal with it. An example of conventional filter,  two-stage csconv filter, and three-stage csconv filter is shown in Fig.~\ref{Fig3}. The input patch is of size $ 7\times 7 $ with $ N_{1} $ channels, the conventional filter in Fig.~\ref{Fig3}(a) is of size $ 7\times 7 $, the two-stage csconv filter in Fig.~\ref{Fig3}(b) is $ [(5\times 5,1\times 1),(3\times 3,1\times 1) ]$, and the three-stage csconv filter in Fig.~\ref{Fig3}(c) is $ 3*(3\times 3,1\times 1) $. In Fig.~\ref{Fig3}(a), the convolution directly generates an output of size $ 1\times 1 $ with $ N_{2} $ channels. In Fig.~\ref{Fig3}(b), feature channels of size $ 3\times 3 $ are obtained by applying the $ (5\times 5,1\times 1) $  subpatch filter, and then an output of size $ 1\times 1 $ with $ N_{2} $ channels is obtained by applying the $ (3\times 3,1\times 1) $  subpatch filter on them. In Fig.~\ref{Fig3}(c), feature channels of size $ 5\times 5 $ are firstly obtained by applying the first $ (3\times 3,1\times 1) $  subpatch filter. And then feature channels of size $ 3\times 3 $ are obtained by applying the second $ (3\times 3,1\times 1) $  subpatch filter on feature channels of size $ 5\times 5 $. Finally, an output of size $ 1\times 1 $ with $ N_{2} $ channels is obtained by applying the third  $ (3\times 3,1\times 1) $ subpatch filter on feature channels of size $ 3\times 3 $. It is seen that the conventional convolution is the most simplest one and that the csconv convolution with a three-stage csconv filter is the most complex one.

\subsection{Form Cascaded Subpatch Network (CSNet) by Stacking Csconv Layers}

\begin{figure*}[!t]
\centering
\includegraphics[scale=0.5]{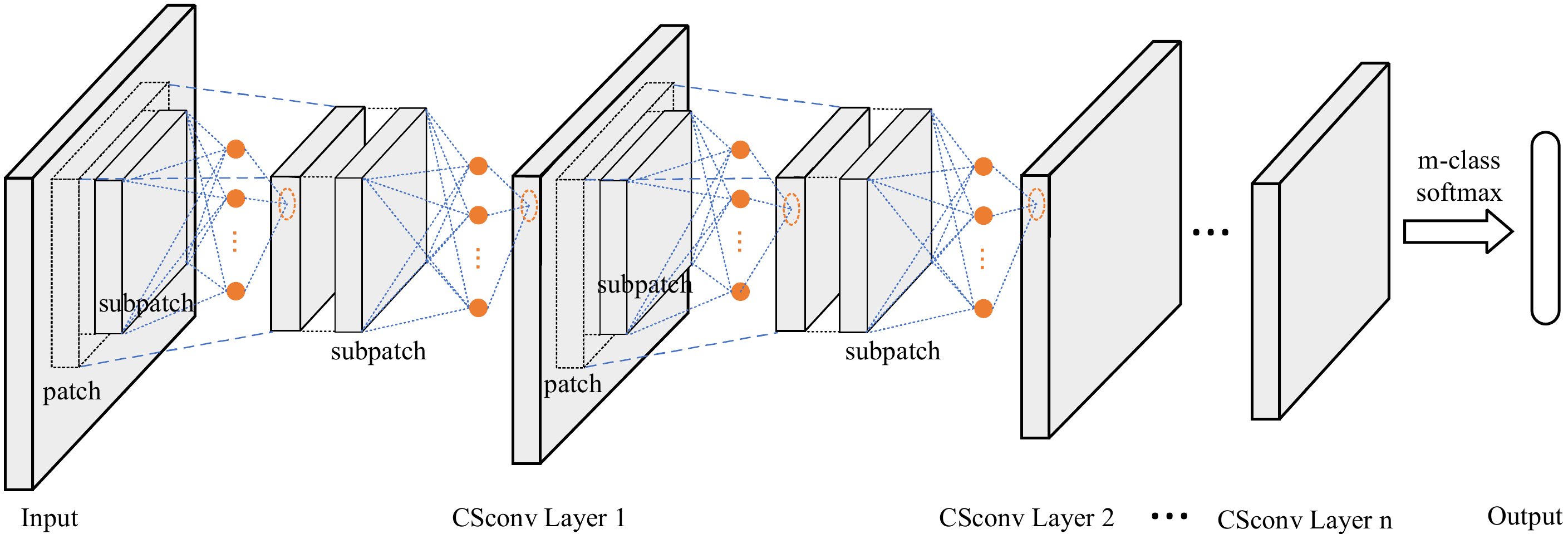}
\caption{The overall structure of CSNet. The number of csconv filters and the number of subpatch filters in each csconv filter can be tuned to deal with different tasks.}
\label{Fig4}
\end{figure*}

Let $ F_{1}= [ (h_{1}^{(1)}\times w_{1}^{(1)},1\times 1),(h_{2}^{(1)}\times w_{2}^{(1)},1\times 1),...,(h_{n}^{(1)}\times w_{n}^{(1)},1\times 1)] $ be a csconv filter. Applying $ F_{1} $ on an input patch yields a $ 1\times 1 $ unit and convolving $ F_{1} $ over the whole input channels yields a convolutional layer $ C_{1} $ (called csconv layer) containing a number of $ 1\times 1 $ units. As the conventional CNN, we can create a new CNN (called CSNet) by stacking a number of csconv layers: $ C_{1},C_{2},...,C_{m} $ with 
$ C_{i}=[ (h_{1}^{(i)}\times w_{1}^{(i)}\times N_{1}^{(i)}\times M_{1}^{(i)},1\times 1\times M_{1}^{(i)}\times Q_{1}^{(i)}),(h_{2}^{(i)}\times w_{2}^{(i)}\times N_{2}^{(i)}\times M_{2}^{(i)},1\times 1\times M_{2}^{(i)}\times Q_{2}^{(i)}),...,(h_{n_{i}}^{(i)}\times w_{n_{i}}^{(i)}\times N_{n_{i}}^{(i)}\times M_{n_{i}}^{(i)},1\times 1\times M_{n_{i}}^{(i)}\times Q_{n_{i}}^{(i)})] $. It is noted that we express both spatial filter and channel filter as four-order tensors by explicitly writing the number of input channels ($ N_{n_{i}}^{(i)}$ for spatial filter and $ M_{n_{i}}^{(i)}$ for channel filter) and the number of output channels ($ M_{n_{i}}^{(i)}$ for spatial filter and $ Q_{n_{i}}^{(i)}$ for channel filter).

It is noted that the number of subpatch filters of a csconv filter $ C_{i} $ is determined by the spatial size of the input patch and the spatial size of each subpatch filter (see Eq.~\ref{eq4}). It is also noted that different csconv layers can have either different or the same configuration of csconv filters. For example, the first three csconv layers of one CSNet can have csconv filters of $ [(5\times 5,1\times 1),(3\times 3,1\times 1)]$ , $ 2*(3\times 3,1\times 1) $, and $ 3*(3\times 3,1\times 1) $, respectively. Fig.~\ref{Fig4} shows the overall structure of the proposed CSNet. The first two csconv layers both have a two-stage csconv filter. The number of csconv filters and the number of subpatch filters in each csconv filter can be tuned according to different tasks. In the proposed CSNet, Rectified Linear Units (ReLUs)\cite{Nair_RectifiedLinear_ICML_2010} follows the output of each convolution of the subpatch filter. Sub-sampling layers can be added in between the csconv layers as in CNN if necessary.

\subsection{Computational Complex Analysis}

Though the csconv convolution is much more complex than conventional convolution, it does not mean that the parameters of one deep CSNet have to be very huge. A comparison of parameters consumed by conventional convolution and the csconv convolution is shown in Tab.~\ref{Tab1}. Suppose that a conventional convolution has a filter of size ${{7\times 7}\times N_{1}}\times N_{2}$ (see Fig.~\ref{Fig3}(a)), where ${7\times 7} $ is the size of the convolutional filter in spatial domain, $ N_{1} $ is the number of input feature channels, and  $ N_{2} $ is the number of output feature channels. The corresponding three-stage csconv convolution ($ 3*(3\times 3,1\times 1) $, see Fig.~\ref{Fig3}(c)) can be implemented with different number of input channels and output channels. Table~\ref{Tab1} presents configurations of two common csconv layers denoted by csconv 1 and csconv 2. As shown in Table~\ref{Tab1}, the parameters consumed by conventional convolution, csconv 1 and csconv 2 are $f_{c}=49N_{1}N_{2}$ 
, $f_{c_{1}}=9N_{1}N_{2}\!+\!21{N_{2}}^{2}$  and $f_{c_{2}}=N_{1}N_{2}+29{N_{1}}^{2}$, respectively. Therefore, the difference between conventional convolution and csconv 1 is $f_{c}-f_{c_{1}}=N_{2}\times (40{N_{1}}-21{N_{2}}) $ . If $ N_{2}\leq\frac{40}{21}N_{1} $, then the number of parameters consumed by csconv 1 is no larger than that of conventional convolution. Similarly, the difference between conventional convolution and csconv 2 is $f_{c}-f_{c_{2}}=N_{1}\times (48{N_{2}}-29{N_{1}}) $. If $ N_{2}\geq\frac{29}{48}N_{1}$, then the number of parameters consumed by csconv 2 is no larger than that of conventional convolution. Especially, if $ N_{1}=N_{2} $, then $ f_{c}= 49{N_{1}}^{2}$, $  f_{c_{1}}= 30{N_{1}}^{2} $, and $  f_{c_{1}}= 30{N_{1}}^{2} $. It is obviously seen that the number of parameters consumed by conventional convolution is lager than that of the proposed  csconv convolution.

In case that $  N_{1} $ and $ N_{2} $ do not satisfy the constraints above, the $ 1\times 1 $ convolution can be used as reduction layer to reduce the number of intermediate output feature channels. This can guarantee that the total number of parameters consumed by csconv convolution is no larger than that of conventional convolution. Since the parameters of each csconv layer are no larger than those of the corresponding conventional convolutional layer, the total parameters of one deep CSNet are also no larger than those of the corresponding conventional neural network. 

\begin{table}[!t]
\renewcommand{\arraystretch}{1.5}
\caption{ The number of parameters consumed by conventional convolution and csconv convolution}
\label{Tab1}
\centering
\begin{tabular}{|c|c|c|c|}
\hline
Method& Conventional& csconv 1& csconv 2 \\
\hline
\multirow{6}*{Structure} & \multirow{6}*{${{7\!\times\! 7}\!\times\! N_{1}}\!\times\! N_{2}$} 
& ${{[(3 \!\times\! 3} \!\times\! N_{1}}\!\times\!  N_{2},$ &${{[(3\!\times\! 3}\!\times\! N_{1}}\!\times\!  N_{1},$\\
& & ${{1\!\times\! 1} \!\times\! N_{2}}\!\times\!  N_{2}),$ &${{1\!\times\! 1}\!\times\! N_{1}}\!\times\! N_{1}),$\\
& & ${{(3\!\times\! 3}\!\times\! N_{2}}\!\times\!  N_{2},$& ${{(3\!\times\! 3}\!\times\! N_{1}}\!\times\! N_{1},$ \\
& & ${{1\!\times\! 1}\!\times\! N_{2}}\!\times\! N_{2}),$ &${{1\!\times\! 1}\!\times\! N_{1}}\!\times\! N_{1}),$\\
& & ${{(3\!\times\! 3}\!\times\! N_{2}}\!\times\! N_{2},$ &${{(3\!\times\! 3}\!\times\! N_{1}}\!\times\! N_{1},$ \\
& & ${{1\!\times\! 1}\!\times\! N_{2}}\!\times\! N_{2})]$ &${{1\!\times\! 1}\!\times\! N_{1}}\!\times\! N_{2})]$\\
\hline
{\#}params & \multirow{2}*{$49N_{1}N_{2}$}
& \multirow{2}*{$9N_{1}N_{2}\!+\!21{N_{2}}^{2}$}  & \multirow{2}*{$N_{1}N_{2}+29{N_{1}}^{2}$}\\
$ N_{1}\!\neq\! N_{2}$& & & \\
\hline
{\#}params & \multirow{2}*{$49{N_{1}}^{2}$}
& \multirow{2}*{$30{N_{1}}^{2}$}  & \multirow{2}*{$30{N_{1}}^{2}$}\\
$ N_{1}\!=\! N_{2}$& & & \\
\hline
\end{tabular}
\end{table}

\section{Experimental Results}

We evaluate the proposed CSNet on four standard benchmark datasets: CIFAR10 \cite{Goodfellow_MulitDigit_CoRR2013}, CIFAR100 \cite{Goodfellow_MulitDigit_CoRR2013}, MNIST \cite{LeCun_Gradient_IEEE1998}, and SVHN \cite{Goodfellow_MulitDigit_CoRR2013}. We compare our CSNets with a dozen well known networks that have achieved the state-of-the-art performance on the four datasets. These networks include Maxout (Maxout Networks) \cite{Goodfellow_Maxout_CoRR2013}, NIN (Network in Network) \cite{Lin_NIN_CoRR}, NIN+LA (Networks with Learned Activation Functions) \cite{Agostinelli_LearningActivation_CoRR2014}, FitNet (Thin and Deep Networks) \cite{Romero_Fitnets_CoRR}, DSN (Deeply Supervised Networks) \cite{Lee_DeeplySupersived_CoRR214}, DropConnect (Networks using Dropconnect) \cite{Wan_RegularizationDropConnect_ICML2013}, dasNet (Deep Attention Selective Networks) \cite{Stollenga_DeepSelective_NIPS2014}, Highway (Networks Allowing Information Flow on Information Highways) \cite{Srivastava_TrainingVeryDeep_NIPS2015}, ALL-CNN (ALL Convolutional Networks) \cite{Springenberg_StrivingSimplicity_CoRR2014}, RCNN (Recurrent Convolutional Neural Networks) \cite{Liang_RecurrentConvolu_CVPR2015}, and  ResNet (Deep Residual Networks) \cite{He_DeepResidual_CoRR2015}.

\subsection{Configuration}
We adopt the global average pooling scheme introduced in \cite{Lin_NIN_CoRR}  on the top layer of CSNet. We also incorporate dropout layers with dropout rate of 0.5 \cite{Goodfellow_Maxout_CoRR2013} . In addition, we use Batch Normalization (BN) \cite{Ioffe_BatchNorm_CorRR}  to accelerate the training stage. The CSNet is implemented using the MatConvNet \cite{Vedaldi_MatConvNet_ACM2015} toolbox in the Matlab environment. We follow a common training protocol \cite{Goodfellow_Maxout_CoRR2013}  in all experiments. We use stochastic gradient descent technique with mini-batch of size 100 at a fixed constant momentum value of 0.9. Initial value for learning rate and weight decay factor is determined based on the validation set. The proposed CSNet is easy to converge and no particular engineering tricks are adopted in all our experiments. All the results are achieved without using the model averaging \cite{Krizhevsky_Imagenet_NIPS2012}  techniques which can help improve the performance.

To comprehensively evaluate the performance of the proposed CSNet, we design three CSNets of different architectures, each of which has different number of parameters. Our small CSNet (CSNet-S), middle CSNet (CSNet-M) and large CSNet (CSNet-L) have 0.96M, 1.6M and 3.5M parameters, respectively. The configurations of CSNet-S, CSNet-M, and CSNet-L are given in Table~\ref{Tab2}. And the corresponding overall structures are presented Fig.\ref{Fig6}. Though the three CSNets are specifically designed for the CIFAR10 dataset, they are also applied on the other three datasets with all the parameters almost remaining the same. The only modification is to change the number of output feature channels of the last csconv layer from 10 to 100 on CIFAR100 dataset.

\begin{figure}[!t]
\centering
\includegraphics[scale=0.6]{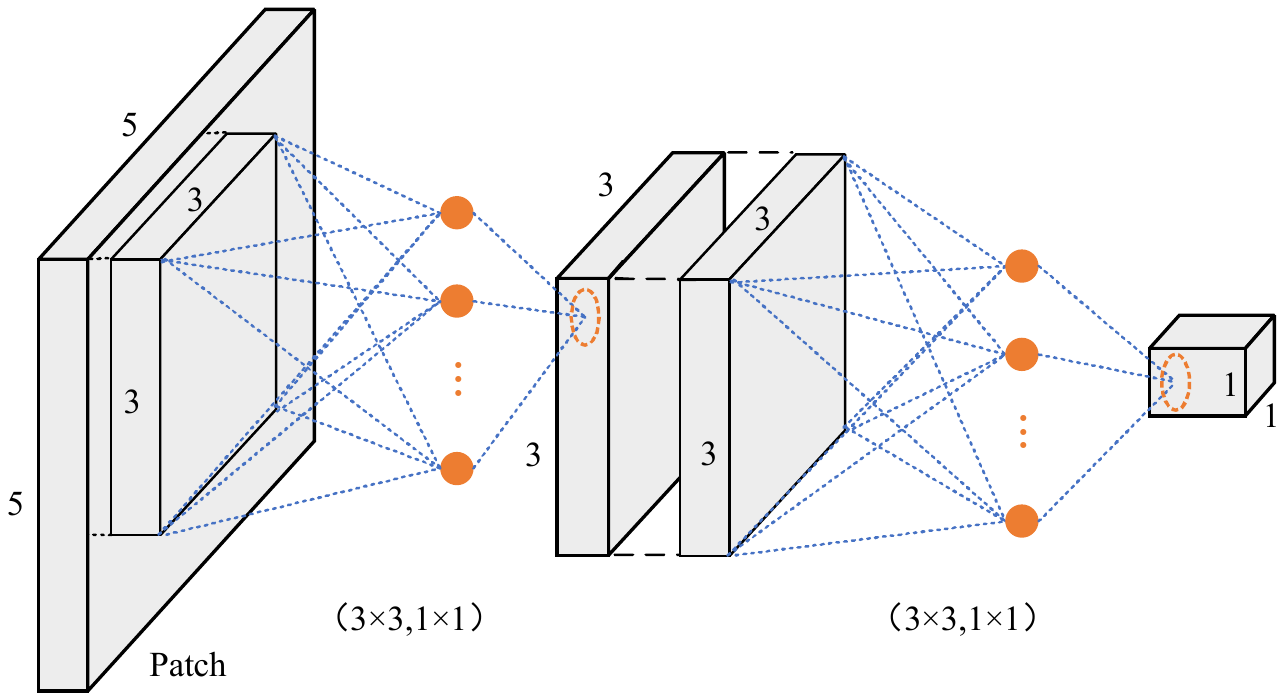}
\caption{Abstracting one patch of size $ 5\times 5 $ with a two-stage csconv filter $2*(3\times 3,1\times 1). $}
\label{Fig5}
\end{figure}

\begin{figure*}[!t]
\centering
\subfloat[CSNet-S]{\includegraphics[scale=0.7]{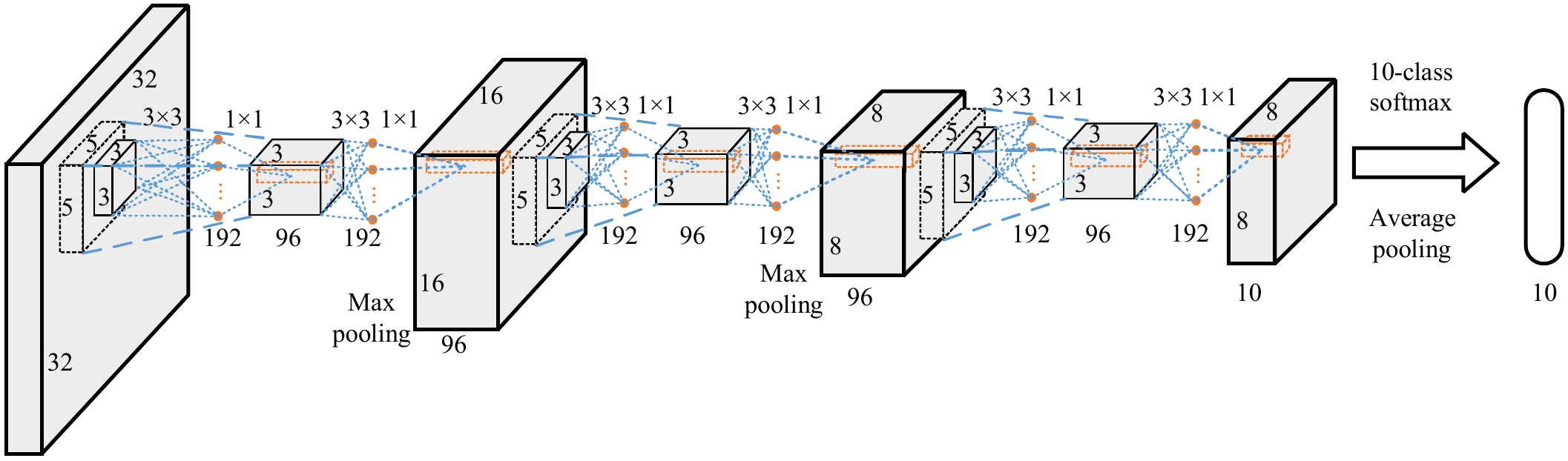}
\label{Fig6(a)}}
\hfil
\subfloat[CSNet-M]{\includegraphics[scale=0.7]{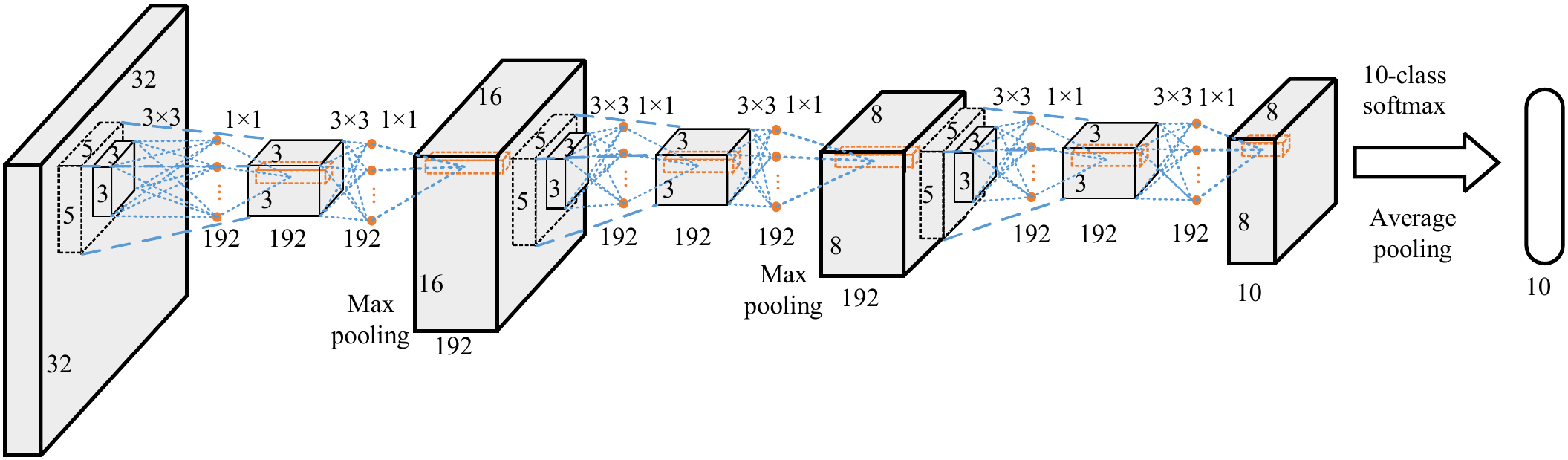}
\label{Fig6(b)}}
\hfil
\subfloat[CSNet-L]{\includegraphics[scale=0.65]{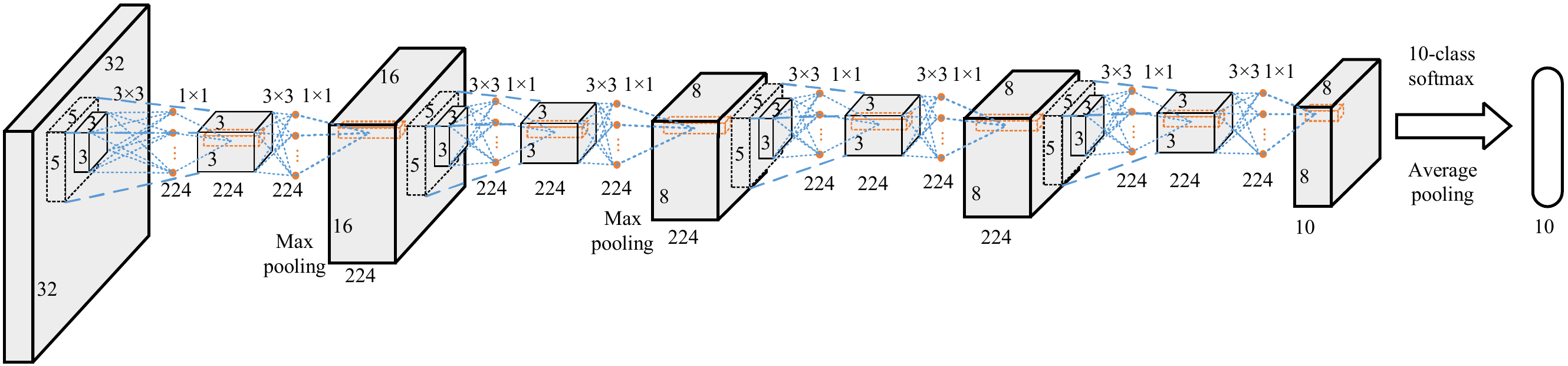}
\label{Fig6(c)}}
\caption{ Overall structures of three different CSNets. The three CSNets are designed for the CIFAR10 dataset. However, they are also applied on the other three datasets with all the parameters almost remaining the same. (a) The CSNet-S with three csconv layers. (b) The CSNet-M with three csconv layers. (c)  The CSNet-L with four csconv layers.}
\label{Fig6}
\end{figure*}

As shown in Table~\ref{Tab2}, the CSNet-S and the CSNet-M have three csconv layers, and the CSNet-L has four csconv layers. Since the input sample is small ($ 32\times 32 $ or $ 28\times 28 $), the receptive field of the filters adopted by traditional methods is typically of size $ 5\times 5$. Therefore, our CSNets use csconv filters of $2*(3\times 3,1\times 1) $ to replace linear filters of size $ 5\times 5$. Fig.~\ref{Fig5} shows the two-stage csconv filter used in our experiment. As shown in Fig.~\ref{Fig6}, max-pooling follows the first two csconv filters of each CSNet. Average-pooling is applied after the last csconv layer to assign one single score for each class. Softmax classifier is then used to recognize the objects. 

\begin{table}[!t]
\renewcommand{\arraystretch}{1.5}
\caption{ Configurations of three different CSNets. Two-stage csconv filters are used to represent patches of size $ 5\times 5 $  }
\label{Tab2}
\centering
\begin{tabular}{|c|c|}
\hline
\multicolumn{2}{|c|}{CSNet-S}  \\
\hline
Patch size& 
5x5 \\
\hline
$ C_{1} $ &$ [(3\!\times\! 3\!\times\! 3\!\times\! 192,1\!\times\! 1\!\times\! 192\!\times\! 96),$\\
& $(3\!\times\! 3\!\times\! 96\!\times\! 192,1\!\times\! 1\!\times\! 192\!\times\! 96)] $\\
$ C_{2} $ &$ [(3\!\times\! 3\!\times\! 96\!\times\! 192,1\!\times\! 1\!\times\! 192\!\times\! 96),$\\
& $(3\!\times\! 3\!\times\! 96\!\times\! 192,1\!\times\! 1\!\times\! 192\!\times\! 96)] $\\
$ C_{3} $ &$ [(3\!\times\! 3\!\times\! 96\!\times\! 192,1\!\times\! 1\!\times\! 192\!\times\! 96),$\\
& $(3\!\times\! 3\!\times\! 96\!\times\! 192,1\!\times\! 1\!\times\! 192\!\times\! 10)] $ \\
\hline
{\#}params& 
0.96M \\
\hline
\multicolumn{2}{|c|}{CSNet-M}  \\
\hline
Patch size& 
5x5 \\
\hline
$ C_{1} $ &$ [(3\!\times\! 3\!\times\! 3\!\times\! 192,1\!\times\! 1\!\times\! 192\!\times\! 192),$\\
& $(3\!\times\! 3\!\times\! 192\!\times\! 192,1\!\times\! 1\!\times\! 192\!\times\! 192)] $\\
$ C_{2} $ &$ [(3\!\times\! 3\!\times\! 192\!\times\! 192,1\!\times\! 1\!\times\! 192\!\times\! 192),$\\
&$(3\!\times\! 3\!\times\! 192\!\times\! 192,1\!\times\! 1\!\times\! 192\!\times\! 192)] $\\
$ C_{3} $ &$ [(3\!\times\! 3\!\times\! 192\!\times\! 192,1\!\times\! 1\!\times\! 192\!\times\! 192),$\\
&$(3\!\times\! 3\!\times\! 192\!\times\! 192,1\!\times\! 1\!\times\! 192\!\times\! 10)] $\\
\hline
{\#}params& 
1.6M \\
\hline
\multicolumn{2}{|c|}{CSNet-L}  \\
\hline
Patch size& 
5x5 \\
\hline
$ C_{1} $ &$ [(3\!\times\! 3\!\times\! 3\!\times\! 224,1\!\times\! 1\!\times\! 224\!\times\! 224),$\\
&$(3\!\times\! 3\!\times\! 224\!\times\! 224,1\!\times\! 1\!\times\! 224\!\times\! 224)] $\\
$ C_{2} $ &$ [(3\!\times\! 3\!\times\! 224\!\times\! 224,1\!\times\! 1\!\times\! 224\!\times\! 224),$\\
&$(3\!\times\! 3\!\times\! 224\!\times\! 224,1\!\times\! 1\!\times\! 224\!\times\! 224)] $\\
$ C_{3} $ &$ [(3\!\times\! 3\!\times\! 224\!\times\! 224,1\!\times\! 1\!\times\! 224\!\times\! 224),$\\
&$(3\!\times\! 3\!\times\! 224\!\times\! 224,1\!\times\! 1\!\times\! 224\!\times\! 224)] $\\
$ C_{4} $ &$ [(3\!\times\! 3\!\times\! 224\!\times\! 224,1\!\times\! 1\!\times\! 224\!\times\! 224),$\\
&$(3\!\times\! 3\!\times\! 224\!\times\! 224,1\!\times\! 1\!\times\! 224\!\times\! 10)] $ \\
\hline
{\#}params& 
3.5M \\
\hline
\end{tabular}
\label{tab1}
\end{table}

\subsection{Results on the CIFAR10 Dataset}

CIFAR10 dataset \cite{Krizhevsky_LearningMultiple_Master2009}  consists of 10 classes of images with 50K training images and 10K testing images. These images are $ 32\times 32 $ color images including airplanes, automobiles, ships, trucks, horses, dogs, cats, birds, deers and frogs. Before training, we preprocess these images using global contrast normalization and ZCA whitening. We carry on experiments with and without data augmentation, respectively. For a fair comparison, we obtain the augmented dataset by padding 4 pixels on each side, and then doing random cropping and random flipping on the fly during training. The augmented data is denoted by CIFAR10$^{+} $. During testing, we only evaluate the single view of the original $ 32\times 32 $ color image.

\begin{table}[!t]
\renewcommand{\arraystretch}{1.5}
\caption{ Quick overview of comparison between (small, middle and large) CSNets and the corresponding counterparts. The results are reported on CIFAR10 in the form of classification error (in \%)}
\label{Tab3}
\centering
\begin{tabular*}{8cm}{@{\extracolsep{\fill}}lcccc}
\hline
Methods & {\#layers} & {\#params} & CIFAR10 &CIFAR10$^{+} $\\
\hline
NIN\cite{Lin_NIN_CoRR}        &9   &0.97M &10.41 &8.81\\
CSNet-S         &12  &0.96M &\textbf{8.33}  &\textbf{6.98}\\
\hline
ResNet-110\cite{He_DeepResidual_CoRR2015}   &110 &1.7M  &-    &6.43\\
CSNet-M         &12  &1.6M  &8.15  &\textbf{6.38}\\
\hline
ResNet-1202\cite{He_DeepResidual_CoRR2015}  &1202 &19.4M  &-  &7.93\\
CSNet-L	         &16 &3.5M &7.74 &\textbf{5.68}\\
\hline	
\end{tabular*}
\end{table}

To have a quick overview of the performance of the CSNets, we firstly compare CSNets with two well known neural networks on this dataset. The first one is the classic NIN network which has 0.97M parameters. The second one is a new network called ResNet \cite{He_DeepResidual_CoRR2015}  which is the champion of the ILSVRC 2015 \cite{Russakovsky_ImagenetLarge_IJCV2015}  classification task. ResNet-110 \cite{He_DeepResidual_CoRR2015}  is a really deep neural network which has up to 110 layers and 1.7M parameters. ResNet-1202 \cite{He_DeepResidual_CoRR2015} is even much deeper and has 19M parameters. It can be seen that the CSNet-S (0.96M) has a little fewer parameters than NIN, and that the CSNet-M (1.6M) has 0.1M fewer parameters than ResNet110, and that the CSNet-L (3.5M) has much fewer parameters than ResNet1202. 

The comparison results are presented in Tab.~\ref{Tab3}. Compared with NIN, the CSNet-S reduces the test error from $ 10.41\% $ to $ 8.33\% $ (without data augmentation), which improves the performance by more than two percent. The CSNet-M obtains a test error of $ 6.38\% $ which is a slightly lower than $ 6.44\% $ of ResNet110 (with data augmentation). However, the CSNet-M has only 12 layers which are much fewer than the 110 layers of ResNet-110. Therefore, it is much easier to train CSNet-M than ResNet-110.  Unlike ResNet-1020 which degrades the performance due to the huge parameters, our CSNet-L further reduces the test error to $ 5.68\% $. The above comparison results demonstrate the superiority of the proposed CSNets. A comprehensive comparison of various methods is presented in Tab.~\ref{Tab4}. It can be seen that the CSNet-S is already among the state-of -the-art results. The CSNet-M surpasses ResNet-100 by $ 0.05\% $ and the CSNet-L surpasses ResNet-100 by $ 0.8\% $.

\begin{table}[!t]
\renewcommand{\arraystretch}{1.5}
\caption{ Classification error (in \%) for CIFAR10 using various methods. A `-' indicates the cited work did not present results for that dataset}
\label{Tab4}
\centering
\begin{tabular*}{6cm}{@{\extracolsep{\fill}}lcc}
\hline
Methods & CIFAR10 & CIFAR10$^{+} $\\
\hline

Maxoutt\cite{Goodfellow_Maxout_CoRR2013}       &11.68 &9.38\\
NIN\cite{Lin_NIN_CoRR}         &10.41 &8.81\\
NIN+LA\cite{Agostinelli_LearningActivation_CoRR2014}      &9.59  &7.51\\
FitNet\cite{Romero_Fitnets_CoRR}       &-    &8.39\\
DSN\cite{Lee_DeeplySupersived_CoRR214}         & 9.75  &8.22\\
DropConnect\cite{Wan_RegularizationDropConnect_ICML2013}  &9.41  &-\\
dasNet\cite{Stollenga_DeepSelective_NIPS2014}       &9.22  &-\\
Highway\cite{Srivastava_TrainingVeryDeep_NIPS2015}      & -    &7.54\\
ALL-CNN\cite{Springenberg_StrivingSimplicity_CoRR2014}     & 9.08 &7.25\\
RCNN-160\cite{Liang_RecurrentConvolu_CVPR2015}     & 8.69 &7.09\\
ResNet-110\cite{He_DeepResidual_CoRR2015}   & -    &6.43 \\
ResNet-1202\cite{He_DeepResidual_CoRR2015}  & -    &7.93\\
\hline
CSNet-S         &8.33  &6.98\\
CSNet-M           &8.15  &6.38\\
CSNet-L	          &\textbf{7.74} &\textbf{5.68}\\
\hline	
\end{tabular*}
\end{table}

\subsection{Results on the CIFAR100 Dataset}
The CIFAR100 dataset \cite{Krizhevsky_LearningMultiple_Master2009}  is just like the CIFAR10 dataset. It has the same amount of training images and testing images as the CIFAR10. However, CIFAR100 contains 100 classes which are ten times of those of CIFAR10. Therefore, the number of images in each class is only one tenth of CIFAR10. The 100 classes in CIFAR100 are grouped into 20 super-classes. Each image has two labels. One is the "fine" label indicating the specific class and the other one is the "coarse" label indicating the super-class. Considering the number of training images for each class, it is much more difficult to recognize the 100 classes of CIFAR100 than the 10 classes of CIFAR10. There is no data augmentation for CIFAR100. We use the same data preprocessing methods as in CIFAR00.

\begin{table}[!t]
\renewcommand{\arraystretch}{1.5}
\caption{ Classification error (in \%) for CIFAR100 using various methods}
\label{Tab5}
\centering
\begin{tabular*}{5cm}{@{\extracolsep{\fill}}lc}
\hline
Methods & CIFAR100\\
\hline
Maxout\cite{Goodfellow_Maxout_CoRR2013}      &38.57\\
NIN\cite{Lin_NIN_CoRR}         &35.68\\
NIN+LA\cite{Agostinelli_LearningActivation_CoRR2014}      &34.40\\
FitNet\cite{Romero_Fitnets_CoRR}      &35.04\\
DSN\cite{Lee_DeeplySupersived_CoRR214}         &34.57\\
dasNet\cite{Stollenga_DeepSelective_NIPS2014}       &33.78\\
ALL-CNN\cite{Springenberg_StrivingSimplicity_CoRR2014}      &33.71\\
Highway\cite{Srivastava_TrainingVeryDeep_NIPS2015}      &32.24\\
RCNN-160\cite{Liang_RecurrentConvolu_CVPR2015}     &31.75\\
\hline
CSNet-M         &\textbf{30.24}\\
\hline	
\end{tabular*}
\end{table}

Since there are 100 classes to be recognized, we adopt the CSNet-M in this experiment. The only difference is that the last convolutional layer of the third csconv layer outputs 100 feature channels, each of which is then averaged to generate one score for one specific class. Details of performance comparison are shown in Tab.~\ref{Tab5}. It can be seen that CSNet-M obtains a test error of $ 30.24\% $ for CIFAR100, which surpasses the second best performance (RCNN-160 with $ 31.75\% $) by 1.51 percent. It also should be noted that RCNN-160 has 1.87M parameters, which are about 0.27M larger than those of CSNet-M.

\subsection{Results on the MNIST Dataset}
MNIST \cite{LeCun_Gradient_IEEE1998}  is one of the most well known datasets in the field of machine learning. It consists of hand written digits ranging from 0 to 9. There are 60000 training images and 10000 testing images which are $ 28\times 28 $ gray-scale images. Only mean subtraction is used to preprocess the dataset. Since MNIST is relatively a simpler dataset compared with CIFAR10, CSNet-S is used in this experiment. The results of performance comparison are shown in Tab.~\ref{Tab6}. It can be seen that CSNet-S achieves the state-of-the-art performance with a test error of $ 0.31\% $. 

\begin{table}[!t]
\renewcommand{\arraystretch}{1.5}
\caption{ Classification error (in \%) for MNIST using various methods}
\label{Tab6}
\centering
\begin{tabular*}{5cm}{@{\extracolsep{\fill}}lc}
\hline
Methods & MNIST\\
\hline
DropConnect\cite{Wan_RegularizationDropConnect_ICML2013}  &0.57 \\
FitNet\cite{Romero_Fitnets_CoRR}      &0.51\\
NIN\cite{Lin_NIN_CoRR}         &0.47\\
Maxout\cite{Goodfellow_Maxout_CoRR2013}      &0.45\\
Highway\cite{Srivastava_TrainingVeryDeep_NIPS2015}     &0.45\\
DSN\cite{Lee_DeeplySupersived_CoRR214}        &0.39\\
RCNN-96\cite{Liang_RecurrentConvolu_CVPR2015}     &\textbf{0.31}\\
\hline
CSNet-S        &\textbf{0.31}\\
\hline	
\end{tabular*}
\end{table}

\subsection{Results on the SVHN Dataset}
The SVHN (Street View House Numbers) \cite{Goodfellow_MulitDigit_CoRR2013}  is a real-world image dataset containing 10 classes representing digits of 0 to 9. There are totally 630,420 $ 32\times32 $ color images which are divided into three sets, 73,527 images in training set, 26,032 images in testing set, and 531,131 images in extra set. More than one digit may exist in an image, and the task is to classify the digit located at the center. We followed the training and testing procedure described by Goodfellow et al. \cite{Goodfellow_Maxout_CoRR2013}.  That is, 400 samples per class are randomly selected from the training set, and 200 samples per class are randomly selected from the extra set. These selected data together form the validation set. The remaining images in the training set and extra set compose the training set. The validation set is only used for tuning hyper-parameter selection and not used during training.
Since there are large variations among one same kind of digit in SVHN due to the changes of color and brightness, it is much more difficult to recognize digits in SVHN than in MNIST. Therefore, local contrast normalization is used to preprocess the samples. No data augmentation is used in this experiment. To deal with the large variations of digits, we use the CSNet-M in this experiment. The performance comparison with other methods is shown in Tab.~\ref{Tab7}. It can be seen that CSNet-M obtains a test error test error of $ 1.9\% $, which already improves NIN ($ 2.35\% $ with 1.98M parameters) by 0.45 percent. CSNet-M achieves the second best performance ($ 1.9\% $) which is very close the best performance with a test error of $ 1.8\% $.

\begin{table}[!t]
\renewcommand{\arraystretch}{1.5}
\caption{ Classification error (in \%) for SVHN  using various methods}
\label{Tab7}
\centering
\begin{tabular*}{5cm}{@{\extracolsep{\fill}}lc}
\hline
Methods & SVHN \\
\hline
Maxout\cite{Goodfellow_Maxout_CoRR2013}      &2.47\\
FitNet\cite{Romero_Fitnets_CoRR}     &2.42\\
NIN\cite{Lin_NIN_CoRR}         &2.35\\
DropConnect\cite{Wan_RegularizationDropConnect_ICML2013}  &1.94 \\
DSN\cite{Lee_DeeplySupersived_CoRR214}         &1.92\\
RCNN-160\cite{Liang_RecurrentConvolu_CVPR2015}     &\textbf{1.80}\\
\hline
CSNet-M        &\textbf{1.90}\\
\hline	
\end{tabular*}
\end{table}

\section{Conclusion}
In this paper, we have presented a novel CNN structure called CSNet. The core of CSNet is to represent a local patch with one neuron which is obtained by using cascaded subpatch filters. The subpatch filter has two characteristics: (1) the spatial size of the subpatch filter is smaller than that $ H\times W $ of the input patch, (2) the subpatch filter consists of an $ h\times w $  (with $  h>1 $ and $ w>1 $) filter followed by a $ 1\times 1 $  filter. The role of cascaded subpatch filters can be considered as representing the input patch using a pyramid with the resolution decreasing from  $ H\times W $ to $ 1\times 1 $ . Due to the large ability of feature representation, the proposed method achieves the state-of-the-art performance.

\ifCLASSOPTIONcaptionsoff
  \newpage
\fi

\end{document}